\documentclass[10 pt]{article}
\usepackage[utf8]{inputenc}
\usepackage{lipsum} 
\usepackage{fullpage}
\usepackage[nodayofweek]{datetime}
\usepackage{amsmath} 
\usepackage{amssymb}  
\usepackage{multirow}
\usepackage{graphicx}
\usepackage{subcaption}
\usepackage{multirow}
\usepackage{makecell}
\usepackage{booktabs}
\usepackage{adjustbox}

\usepackage{scalerel,stackengine}
\stackMath
\newcommand\reallywidehat[1]{%
\savestack{\tmpbox}{\stretchto{%
  \scaleto{%
    \scalerel*[\widthof{\ensuremath{#1}}]{\kern-.6pt\bigwedge\kern-.6pt}%
    {\rule[-\textheight/2]{1ex}{\textheight}}
  }{\textheight}%
}{0.5ex}}%
\stackon[1pt]{#1}{\tmpbox}%
}

\makeatletter
\def\thanks#1{\protected@xdef\@thanks{\@thanks
		\protect\footnotetext{#1}}}
\makeatother

\title{\LARGE \bf
	Observability Analysis of Visual-Inertial Odometry with Online Calibration of Velocity-Control Based Kinematic Motion Models
}
\author{Haolong Li$^{1}$ and Joerg Stueckler$^{1}$
		\thanks{$^{1}$ All authors are with the Embodied Vision Group, 
		Max Planck Institute for Intelligent Systems, T\"ubingen, Germany
		{\tt\small \{haolong.li,joerg.stueckler\}@tue.mpg.de}}%
}

\date{}

\begin{document}
\maketitle

\begin{abstract}
	
	In this paper, we analyze the observability of the visual-inertial odometry (VIO) using stereo cameras with a velocity-control based kinematic motion model~\cite{kinvio_arxiv}.
	Previous work shows that in general case the global position and yaw are unobservable in VIO system, additionally the roll and pitch become also unobservable if there is no rotation. We prove that by integrating a planar motion constraint roll and pitch become observable. We also show that the parameters of the motion model are observable. 
	
\end{abstract}

\section{Visual-Inertial Odometry with Velocity-Control Based Kinematic Motion Models}
In this section we give an overview of the VIO with kinematic motion model constraints (Kin-VIO). The Kin-VIO~\cite{kinvio_arxiv} is an extension of the non-linear optimization-based VIO, which uses motion model constraints in addition to visual and IMU measurements to estimate the camera poses. More formally, for each frame at time $t$, the VIO estimates the camera pose $\mathbf{T}_t \in SE(3)$, the camera velocity $\mathbf{v}_t$, and the bias parameters $\mathbf{b}_t$ of the IMU. The state variable $\mathbf{x}$ is:
\begin{equation}
{\mathbf{x}} = \begin{bmatrix}
\mathbf{T} & \mathbf{v} & \mathbf{b}
\end{bmatrix}^{\top}.
\end{equation}
In~\cite{hesch2014_vio_analysis} the authors show that the VIO has four unobservable directions which are the three global translation and one global yaw direction. In~\cite{wu2017_vinsonwheels}, it was shown that the scale is unobservable if the camera has constant acceleration, the roll and pitch become also unobservable if there is no rotation. 

In Kin-VIO we incorporate the velocity-control based motion model~\cite{thrun2005probabilistic} into the
visual-inertial odometry framework. We use stereo camera which makes the scale observable.
Assuming that the robot can be controlled by a control command $\mathbf{u} = ( v, \omega )^\top$ through a linear velocity $v \in \mathbb{R}$ in forward direction and a rotational velocity $\omega \in \mathbb{R}$ along the vertical axis.
The motion model propagates the robot pose $\mathbf{P}_{t} \in SE(2)$ on the ground plane with the control command,
$\mathbf{P}_{t'} =  \mathbf{P}_{t} \exp\left( \Delta t \, \widehat{\boldsymbol{\xi}}  \right)$,
where $\boldsymbol{\xi} = ( v, 0, \omega )$ is the twist vector, $\widehat{\boldsymbol{\xi}} = \left( \begin{array}{ccc} 0 & -\omega & v\\ \omega & 0 & 0\\ 0& 0& 0 \end{array} \right)$ maps a 3D vector to $se(2)$.
Velocity commands are executed by the robot in the robot base frame whose pose relative to the world frame is denoted by $\mathbf{T}_b^w \in SE(3)$ (transforming coordinates from base $b$ to world frame $w$).
In the base frame, the $x$-axis points in forward driving direction, while the $z$-axis points upwards and is the axis of rotational robot motion.
The VIO provides pose estimates of the body frame of the IMU-camera sensor in the world frame, i.e. $\mathbf{T}_i^w \in SE(3)$.
The sensor is placed rigidly on the robot at a relative pose $\mathbf{T}_i^b \in SE(3)$ to the robot base frame.
To quantify the relative motion $\mathbf{T}_{b,t'}^{b,t}$ of the robot base frame from times $t$ to $t'$ of subsequent image frames, we can hence determine
$\mathbf{T}_{b,t'}^{b,t} =  \mathbf{T}_{i,t}^{b,t} \left(\mathbf{T}_{i,t}^{w}\right)^{-1} {\mathbf{T}_{i,t'}^{w}} \left(\mathbf{T}_{i,t'}^{b,t'}\right)^{-1}$.
The rotation $\Delta \theta$ around the $z$-axis of the base frame is calculated from the relative rotation $\mathbf{R}_{b,t'}^{b,t}$ in $\mathbf{T}_{b,t'}^{b,t}$ as the $z$-component of $\log\left( \mathbf{R}_{b,t'}^{b,t} \right)$.
The translational motion $( \Delta x, \Delta y )^\top$ in the $x$-$y$-plane is determined from the corresponding entries of $\mathbf{T}_{b,t'}^{b,t}$. 
The estimated twist is 
\begin{equation}
\boldsymbol{\zeta} = \\ \frac{1}{\Delta t} \log  \left( \begin{array}{ccc}
\cos(\Delta \theta) & -\sin(\Delta \theta) & \Delta x\\
\sin(\Delta \theta) & \cos(\Delta \theta) & \Delta y\\
0 & 0 & 1
\end{array} \right),
\end{equation}
where $\Delta t = t' - t$.
We add residuals of the form
$\mathbf{r}_{\boldsymbol{\xi}} = \boldsymbol{\zeta} - \boldsymbol{\bar{\xi}}$
which implicitly measure the difference between the state estimates and the motion model prediction. 

In practice, the real action of the robots differs from
the received control commands due to effects such as
hardware acceleration limits, time offsets and properties of low-level controllers. To mitigate this difference and build a meaningful residual, we estimate an effective control $\boldsymbol{\bar{\xi}}_t$ at arbitrary time $t$, e.g. at the time of an image frame, from a window of most recent commands.
We average a window of recent control commands with weights determined by a RBF kernel for the translational and rotational parts separately:
\begin{equation}
\boldsymbol{\bar{\xi}}_t = \\ \left(
\begin{array}{c}
s_{lin} \frac{\sum_{\tau \in \mathcal{W}_t} \exp\left( - \frac{\left\| d\tau - \mu_{lin} \right\|^2}{2 \sigma_{lin} ^2} \right) v_{\tau}} {\sum_{\tau \in \mathcal{W}_t} \exp\left( - \frac{\left\| d\tau - \mu_{lin} \right\|^2}{2 \sigma_{lin} ^2} \right)}\\
0 \\
s_{ang} \frac {\sum_{\tau \in \mathcal{W}_t} \exp\left( - \frac{\left\| d\tau - \mu_{ang} \right\|^2}{2 \sigma_{ang} ^2} \right) w_{\tau}} {\sum_{\tau \in \mathcal{W}_t} \exp\left( - \frac{\left\| d\tau - \mu_{ang} \right\|^2}{2 \sigma_{ang} ^2} \right)}
\end{array} \right) .
\end{equation}
Here $\mathcal{W}_t$ is a window of $N$ control commands indexed by their times $\tau$ at or before time $t$ and $d\tau := t - \tau$. 
For an image frame at time $t$, the window typically spans the $N$ control commands that have occurred before the frame.
We optimize for $\mu$ and $\sigma$ and scale factor $s$ of both linear and angular parts as global parameters together with the VIO states. 
The RBF parameters are summarized in the state variables $\mathbf{p}_{\mathit{rbf},t}$ at time $t$.

We exploit prior knowledge that our robot moves on flat ground and can only rotate along the vertical axis of the ground in indoor environments and add a stochastic
plane constraint~\cite{wu2017_vinsonwheels} for the robot pose. 
The plane can be parameterized as a 2 degree-of-freedom quaternion $\mathbf{q}^g_w$ and a scalar $d^g_w$ which represents the distance between the ground plane to the world frame origin. 
The residual is
\begin{equation}
\mathbf{r}_{\mathbf{p}} =  \left( \begin{array}{c}
\left(\mathbf{R}(\mathbf{q}^g_w) \mathbf{R}^w_i (\mathbf{R}^b_i)^\top \mathbf{e}_3 \right)_{1,2}\\
d^g_w + \mathbf{e}_3^\top \mathbf{R}(\mathbf{q}^g_w) (\mathbf{t}^w_i - \mathbf{R}^w_i {\mathbf{R}^b_i}^\top \mathbf{t}^b_i)
\end{array} \right),
\end{equation} 
with $\mathbf{e}_3 = \begin{pmatrix}0 & 0 & 1\end{pmatrix}^\top$.

\section{Observability Analysis of Kin-VIO}
As in~\cite{wu2017_vinsonwheels}, the observability of the state variables can be analyzed based on the underlying state-space model irrespective of the implementation of the estimator.
We follow the proof scheme in~\cite{wu2017_vinsonwheels}, and discuss about the observability of the state variables in this section.

In Kin-VIO the motion constraint and the planar constraint are treated as observation, the augmented state vector includes plane parameters $\mathbf{q}^{g}_{w}$, $d^{g}_{w}$ between ground $g$ and world frame $w$, extrinsic parameters $\mathbf{q}^{b}_i$, $\mathbf{t}^{b}_i$ between robot base and IMU frame and RBF parameters $\mathbf{p}_{\mathit{rbf}}$:
\begin{equation}
\bar{\mathbf{x}} = \begin{bmatrix}
\mathbf{x}^{\top} & \mathbf{q}^{g}_{w} &d^{g}_{w} & \mathbf{q}^{b}_i & \mathbf{t}^{b}_i & \mathbf{p}_{\mathit{rbf}}
\end{bmatrix}^{\top}.
\end{equation}
As in~\cite{wu2017_vinsonwheels}, the state transition is given by the IMU propagation model. The linearized transition matrix $\mathbf{\Phi}_{k,1}$  from time-step 1 to $k$  for the IMU propagation is derived in~\cite{hesch2014_vio_analysis}.
By including the transition of the parameters as a constant propagation model, the transition matrix of our model becomes:
\begin{equation}
\bar{\boldsymbol{\Phi}}_{k,1} = \begin{bmatrix}
\mathbf{\Phi}_{k,1} & \mathbf{0}\\
\mathbf{0} & \mathbf{I}
\end{bmatrix}
\end{equation}
Note that we use a stereo camera in above mentioned method, the scale becomes directly observable and is not discussed in this paper.

\subsubsection{Observability of VIO States and Extrinsic Parameters}
The observability of VIO state and extrinsic parameters are studied in the previous work~\cite{lee2020_viwo_onlinecalib}, where a velocity based forward kinematic motion model is integrated into the VIO. In our Kin-VIO we use an inverse kinematic motion model, here we show that the observability of integrating the inverse kinematic model is the same as using the forward kinematic model.

Given the relative pose between two frames $
\mathbf{T}_{b,t'}^{b,t}$, the corresponding $SE(2)$ pose $\mathbf{P}_{b,t'}^{b,t}$ in the horizontal plane consisting of rotational part $\mathbf{Q}_{b,t'}^{b,t}$ and translational part $\mathbf{p}_{b,t'}^{b,t}$,
the effective control input $\boldsymbol{\xi} = (v, 0, w)^\top$ with linear and angular velocity, and $\bar{\mathbf{P}}^{b,t}_{b,t'} = exp(\Delta t \boldsymbol{\xi})$, the forward model and its derivatives can be written as
\begin{equation}
\begin{split}
\mathbf{r}_{\mathit{forward}} &= f({\mathbf{P}_{b,t'}^{b,t}} , \bar{\mathbf{P}}^{b,t}_{b,t'}) =
\begin{bmatrix}
\mathbf{p}_{b,t'}^{b,t} - \bar{\mathbf{p}}_{b,t'}^{b,t}\\
log_{\mathit{so2}}(\mathbf{Q}_{b,t'}^{b,t}) - w \Delta t
\end{bmatrix} \\
\frac{\partial \mathbf{r}_{\mathit{forward}}}{\partial \mathbf{P}_{b,t'}^{b,t}} &= \frac{\partial f(.)}{\partial {\mathbf{P}_{b,t'}^{b,t}}},
\end{split}
\end{equation}
while the inverse model and the derivatives are
\begin{equation}
\begin{split}
\mathbf{r}_{\mathit{inverse}} &= log_{\mathit{se2}}({\mathbf{P}_{b,t'}^{b,t}}) - \Delta t \boldsymbol{\xi}
= \begin{bmatrix}
log_{\mathit{se2}}({\mathbf{P}_{b,t'}^{b,t}})_{x,y} - v \Delta t\\
log_{\mathit{so2}}(\mathbf{Q}_{b,t'}^{b,t}) - w \Delta t
\end{bmatrix}   \\
\frac{\partial \mathbf{r}_{\mathit{inverse}}}{\partial \mathbf{P}_{b,t'}^{b,t}} &= \frac{\partial log_{\mathit{se2}}(.)}{\partial \mathbf{P}_{b,t'}^{b,t}}.
\end{split}
\end{equation}
While in the forward model the derivative wrt. $\mathbf{p}_{b,t'}^{b,t}$ is an identity matrix, the derivative in the inverse model is $J_{\log_{\mathit{se2}}}$.
Because $J_{\log_{\mathit{se2}}}$ is an invertible matrix by definition,
when we compute the observability matrix by multiplying this Jacobian matrix and the transition matrix, 
the rank of the observability matrix remains the same based on the Sylvester’s inequality. As shown in~\cite{lee2020_viwo_onlinecalib} the extrinsic parameters will be unobservable under certain motions. 
We used the extrinsics derived from the CAD model as an initial Gaussian prior to counteract this problem.

\subsubsection{Observability of Global Orientation and Plane Parameters}
In this subsection we show that the global orientation becomes observable by using the plane constraint and a set of priors in the initial frames.

The rotational part of the pose $\mathbf{T}$ is represented by the quaternion $\mathbf{q}$ and the translational part is $\mathbf{t}$. For simplicity we denote $\mathbf{R}(\mathbf{q}^g_w)$ as $\mathbf{R}^g_w$. The derivative of the rotation part is computed on $so3$.
For the plane constraint, the Jacobians for step $k$ are:
\begin{equation}
\begin{split}
\frac{\partial \mathbf{r}^{1}_{\mathbf{p}}}{\partial \mathbf{q}^{i_k}_{w}}	&=	\mathbf{H}^{1}_{\mathbf{q}^{i_k}_{w}} = (\reallywidehat{\mathbf{R}^g_w \mathbf{R}^w_i (\mathbf{R}^b_i)^\top \mathbf{e}_3} \mathbf{R}^g_w \mathbf{R}^w_i)_{(:2,:)}\\ 
\frac{\partial \mathbf{r}^{1}_{\mathbf{p}}}{\partial \mathbf{q}^{g}_{w}}	&=
\mathbf{H}^{1}_{\mathbf{q}^{g}_{w}} = -(\reallywidehat{\mathbf{R}^g_w \mathbf{R}^w_i (\mathbf{R}^b_i)^\top \mathbf{e}_3})_{(:2,:2)}\\
\frac{\partial \mathbf{r}^{2}_{\mathbf{p}}}{\partial \mathbf{q}^{i_k}_{w}}	&=
\mathbf{H}^{2}_{\mathbf{q}^{i_k}_{w}} =-{\mathbf{e}_{3}^{T}} \mathbf{R}^{g}_w \mathbf{R}^w_{i_k} \widehat{(\mathbf{R}^{i_k}_b \mathbf{t}^b_i) }\\ 
\frac{\partial \mathbf{r}^{2}_{\mathbf{p}}}{\partial \mathbf{q}^{g}_{w}}	&=
\mathbf{H}^{2}_{\mathbf{q}^{g}_{w}} =-{{\mathbf{e}_{3}^{T}} \reallywidehat{(\mathbf{R}^g_w (\mathbf{t}^w_{i_k} - \mathbf{R}^w_{i_k} \mathbf{R}^{i_k}_b \mathbf{t}^{b}_i ))}}_{(:,:2)}\\ 
\frac{\partial \mathbf{r}^{2}_{\mathbf{p}}}{\partial \mathbf{t}^w_{i_k}} &=
\mathbf{H}^{2}_{\mathbf{t}^w_{i_k}} ={\mathbf{e}_{3}^{T}}\mathbf{R}^g_w \\
\frac{\partial \mathbf{r}^{2}_{\mathbf{p}}}{\partial {d}^g_{w}}	&= \quad\mathbf{H}^{2}_{d^g_w} = 1,
\end{split}
\end{equation}
where $\mathbf{r}^{1}_{\mathbf{p}}$ and $\mathbf{r}^{2}_{\mathbf{p}}$ are the first and second row of the plane constraint residual, $\mathbf{H}^1$ and $\mathbf{H}^2$ are the first and second row in the Jacobian matrix. $(.)_{(:2,:)}$ denotes the first two rows and $(.)_{(:,:2)}$ denotes the first two columns.
The observability matrix can be written as:
\begin{equation}
\begin{split}
\mathbf{M}_k^{plane}  &= \mathbf{H}_k^{plane} \bar{\boldsymbol{\Phi}}_{k,1} \\
&= \begin{bmatrix}
\mathbf{H}^{1}_{\mathbf{q}^{i_k}_{w}} &
\dots & \mathbf{0} & \mathbf{H}^{1}_{\mathbf{q}^{g}_{w}} & 0 & \dots\\
\mathbf{H}^{2}_{\mathbf{q}^{i_k}_{w}} & \dots & \mathbf{H}^{2}_{\mathbf{t}^w_{i_k}} & \mathbf{H}^{2}_{\mathbf{q}^{g}_{w}} & 1 & \dots
\end{bmatrix} \bar{\boldsymbol{\Phi}}_{k,1}\\
&=  \begin{bmatrix}
\mathbf{H}^{1}_{\mathbf{q}^{i_k}_{w}} \boldsymbol{\Phi}^{1,1}_{k,1} &
\mathbf{H}^{1}_{\mathbf{q}^{i_k}_{w}} \boldsymbol{\Phi}^{1,2}_{k,1} &\dots & &
& \mathbf{0} & \mathbf{H}^{1}_{\mathbf{q}^{g}_{w}} & 0 & \dots\\
\mathbf{H}^{2}_{\mathbf{q}^{i_k}_{w}} \boldsymbol{\Phi}^{1,1}_{k,1} +
\mathbf{H}^{2}_{\mathbf{t}^w_{i_k}} \boldsymbol{\Phi}^{5,1}_{k,1} &
\mathbf{H}^{2}_{\mathbf{q}^{i_k}_{w}} \boldsymbol{\Phi}^{1,2}_{k,1} +
\mathbf{H}^{2}_{\mathbf{t}^w_{i_k}} \boldsymbol{\Phi}^{5,2}_{k,1} &
\mathbf{H}^{2}_{\mathbf{t}^w_{i_k}} \delta t_{k} &
\mathbf{H}^{2}_{\mathbf{t}^w_{i_k}} \boldsymbol{\Phi}^{5,4}_{k,1} &
\mathbf{H}^{2}_{\mathbf{t}^w_{i_k}} &
\mathbf{0}& \mathbf{H}^{2}_{\mathbf{q}^{g}_{w}} & 1 & \dots
\end{bmatrix}.
\end{split}	
\end{equation}
The derivation of the nullspace $\mathbf{N}_o$ for the global orientation of VIO is given in~\cite{WuReport}. Since the change of the global orientation also affects the plane angle $\mathbf{q}^g_w$, following appendix D in~\cite{WuReport}, the null space of the global orientation in Kin-VIO becomes:
\begin{equation}
\begin{split}
\bar{\mathbf{N}}_o = 
\begin{bmatrix}
\mathbf{N}_o^{\top} & {\mathbf{R}^w_g}_{(:,:2)} & \mathbf{0} & \dots & 
\mathbf{0}
\end{bmatrix}^{\top},
\end{split}
\end{equation}
the product is:
\begin{equation}
\begin{split}
\mathbf{M}_k^{plane}  \bar{\mathbf{N}}_o  &= 
\begin{bmatrix}
(\reallywidehat{\mathbf{R}^g_w \mathbf{R}^w_{ik} (\mathbf{R}^b_i)^\top \mathbf{e}_3} \mathbf{R}^g_w \mathbf{R}^w_{ik})_{(:2,:)} \mathbf{R}^{i_k}_{i_0} \mathbf{R}^{i_0}_w - (\reallywidehat{\mathbf{R}^g_w \mathbf{R}^w_{ik} (\mathbf{R}^b_i)^\top \mathbf{e}_3})_{(:2,:2)} {\mathbf{R}^g_w}_{(:2,:)}\\
-{\mathbf{e}_{3}^{T}} \mathbf{R}^{g}_w \mathbf{R}^w_{i_k} \reallywidehat{(\mathbf{R}^{i_k}_b \mathbf{t}^b_i) }
\mathbf{R}^{i_k}_{i_0} \mathbf{R}^{i_0}_w + {\mathbf{e}_{3}^{T}}\mathbf{R}^g_w
( -\reallywidehat{(\mathbf{t}^w_{i_0} + \mathbf{v}^{w}_{i_0}\delta t_k + \frac{1}{2}\mathbf{g}^w\delta t^2_k - \mathbf{t}^w_{i_k})}  \mathbf{R}^w_{i_0} \mathbf{R}^{i_0}_w +  \delta t_k \reallywidehat{\mathbf{v}^w_{i_0}}  \dots\\
\dots - \boldsymbol{\Phi}^{5,4}_{k,1} {\mathbf{R}^{i_0}_w} \reallywidehat{\mathbf{g}^w}  + \reallywidehat{\mathbf{t}^w_{i_0}} ) {-{{\mathbf{e}_{3}^{T}} \reallywidehat{(\mathbf{R}^g_w (\mathbf{t}^w_{i_k} - \mathbf{R}^w_{i_k} \mathbf{R}^{i_k}_b \mathbf{t}^{b}_i ))}}_{(:,:2)}} {\mathbf{R}^g_w}_{(:2,:)}
\end{bmatrix} \\
&\stackrel{(36) in~\cite{wu2017_vinsonwheels}}{=} \begin{bmatrix}
(\reallywidehat{\mathbf{R}^g_w \mathbf{R}^w_{ik} (\mathbf{R}^b_i)^\top \mathbf{e}_3} \mathbf{R}^g_w)_{(:2,:)} - (\reallywidehat{\mathbf{R}^g_w \mathbf{R}^w_{ik} (\mathbf{R}^b_i)^\top \mathbf{e}_3})_{(:2,:2)} {\mathbf{R}^g_w}_{(:2,:)} \\
-{\mathbf{e}_{3}^{T}} \mathbf{R}^{g}_w \mathbf{R}^w_{i_k} \reallywidehat{(\mathbf{R}^{i_k}_b \mathbf{t}^b_i) } \mathbf{R}^{i_k}_w
+{\mathbf{e}_{3}^{T}} {\mathbf{R}}^g_w \reallywidehat{\mathbf{t}^w_{i_k}} {-{{\mathbf{e}_{3}^{T}} \reallywidehat{(\mathbf{R}^g_w (\mathbf{t}^w_{i_k} - \mathbf{R}^w_{i_k} \mathbf{R}^{i_k}_b \mathbf{t}^{b}_i ))}}_{(:,:2)}} {\mathbf{R}^g_w}_{(:2,:)}
\end{bmatrix} \\
&=\begin{bmatrix}
\mathbf{0}\\
\mathbf{0}
\end{bmatrix},
\end{split}
\end{equation}
with the property of the rotation matrix: $\widehat{\mathbf{R} \mathbf{t}} = \mathbf{R}
\widehat{\mathbf{t}} \mathbf{R}^{T}$. $\delta t_{k}$ in above equation denotes the time difference between time-steps 1 and $k$ and $\mathbf{g}^w$ is the gravity vector in world frame.
Since our robot starts from still state, we use the initial accelerator measurement to initialize the plane angle. 
By adding this information as a Gaussian initial prior on the plane angle $\mathbf{q}^g_{w}$, the global orientation becomes observable. The change of the plane distance $d^g_w$ does not affect other state variables, thus the nullspace for $d^g_w$ is $\left[ \mathbf{0 } \quad 1 \quad \mathbf{0}  \right]$, 
it is observable because $\mathbf{M}^{plane}_k \left[ \mathbf{0 } \quad 1 \quad \mathbf{0}  \right] = \left[0 \quad 1\right]^{\top}$. 

\subsubsection{Observability of RBF Parameters}
Finally, we prove the observability of the RBF parameters ($s_{lin}, \mu_{lin}, \sigma_{lin}$) for the translational velocity. The parameters for the angular velocity have the same observability. We denote $\exp\left( - \frac{\| dt_i - \mu_{lin} \|^2}{2 \sigma_{lin} ^2} \right)$ as $\exp(.) $, where $dt_i$ is the time difference between control command at $t_i$ and image frame at $t$.
The Jacobian matrix is:
\begin{equation}
\begin{split}
\mathbf{H}_{rbf} &= \begin{bmatrix}
\dots & \frac{\partial \mathbf{r}_{\boldsymbol{\xi}}}{\partial s_{lin}} & 
\frac{\partial \mathbf{r}_{\boldsymbol{\xi}}}{\partial \mu_{lin}} & 
\frac{\partial \mathbf{r}_{\boldsymbol{\xi}}}{\partial \sigma_{lin}}
\end{bmatrix} \\
&= \begin{bmatrix}
\dots & -\frac{\bar{v}}{s_{lin}} & \frac{\bar{v} {\sum_{i = 1}^{N} \exp(.)} \frac{dt_i - \mu}{\sigma^2} - s_{lin}  {\sum_{i = 1}^{N} v_i \exp(.)} \frac{dt_i - \mu}{\sigma^2}}{{\sum_{i = 1}^{N} \exp(.)}} &
\frac{\bar{v} {\sum_{i = 1}^{N} \exp(.)} \frac{(dt_i - \mu)^2}{\sigma^3} - s_{lin}  {\sum_{i = 1}^{N} v_i \exp(.)} \frac{(dt_i - \mu)^2}{\sigma^3}}{{\sum_{i = 1}^{N} \exp(.)}}
\end{bmatrix},
\end{split}
\end{equation}
where
\begin{equation}
\bar{v} = 	s_{lin} \frac{\sum_{i = 1}^{N} \exp(.) v_{t_i}} {\sum_{i = 1}^{N} \exp(.)}.
\end{equation}
Following the process in~\cite{WuReport}, we can perturb the state vector $\bar{\mathbf{x}}$ along the unobservable direction of the RBF parameters. Since the change of the RBF parameters does not affect other state variables, the nullspace for the RBF parameters $\mathbf{N}_{\mathit{rbf}}$ is:
\begin{equation}
\mathbf{N}_{rbf} = \begin{bmatrix}
\mathbf{0} & \dots & \mathbf{0} &\mathbf{I}_{3\times 3}
\end{bmatrix}^\top,
\end{equation}
The corresponding $3 \times 3$ bottom right block of the observability matrix for our inverse motion model is:
\begin{equation}
\mathbf{M}_{\mathit{rbf}} = \begin{bmatrix}
-\frac{\bar{v}}{s_{lin}} & 0 & 0 \\
0 & \frac{\bar{v} {\sum_{i = 1}^{N} \exp(.)} \frac{dt_i - \mu_{lin}}{\sigma_{lin}^2} - s_{lin}  {\sum_{i = 1}^{N} v_i \exp(.)} \frac{dt_i - \mu_{lin}}{\sigma_{lin}^2}}{{\sum_{i = 1}^{N} \exp(.)}}& 0 \\
0 & 0 & 	\frac{\bar{v} {\sum_{i = 1}^{N} \exp(.)} \frac{(dt_i - \mu_{lin})^2}{\sigma_{lin}^3} - s_{lin}  {\sum_{i = 1}^{N} v_i \exp(.)} \frac{(dt_i - \mu_{lin})^2}{\sigma_{lin}^3}}{{\sum_{i = 1}^{N} \exp(.)}}
\end{bmatrix}.
\end{equation}
The product of the motion model observability matrix and the nullspace $\mathbf{N}_{\mathit{rbf}}$ is equal to $\mathbf{M}_{\mathit{rbf}}$ as the other terms cancel out by multiplication with the $\mathbf{0}$ components in $\mathbf{N}_{\mathit{rbf}}$.
The RBF parameters are unobservable if $v_i$ in the window are constant. We used their initial values as a weak prior to guarantee that their values don't jump at the beginning of the optimization. Due to the marginalization prior, the RBF parameters will remain observable even if they become temporarily unobservable in the optimization window during the process. 
\bibliographystyle{IEEEtran}
\bibliography{egbib}

\begin{thebibliography}{1}
\providecommand{\url}[1]{#1}
\csname url@samestyle\endcsname
\providecommand{\newblock}{\relax}
\providecommand{\bibinfo}[2]{#2}
\providecommand{\BIBentrySTDinterwordspacing}{\spaceskip=0pt\relax}
\providecommand{\BIBentryALTinterwordstretchfactor}{4}
\providecommand{\BIBentryALTinterwordspacing}{\spaceskip=\fontdimen2\font plus
\BIBentryALTinterwordstretchfactor\fontdimen3\font minus
  \fontdimen4\font\relax}
\providecommand{\BIBforeignlanguage}[2]{{%
\expandafter\ifx\csname l@#1\endcsname\relax
\typeout{** WARNING: IEEEtran.bst: No hyphenation pattern has been}%
\typeout{** loaded for the language `#1'. Using the pattern for}%
\typeout{** the default language instead.}%
\else
\language=\csname l@#1\endcsname
\fi
#2}}
\providecommand{\BIBdecl}{\relax}
\BIBdecl

\bibitem{kinvio_arxiv}
H.~Li and J.~Stueckler, ``Visual-inertial odometry with online calibration of
  velocity-control based kinematic motion models,'' \emph{Accepted for
  publication in Robotics and Automation Letters (RA-L)}, 2022.

\bibitem{hesch2014_vio_analysis}
J.~A. Hesch, D.~G. Kottas, S.~L. Bowman, and S.~I. Roumeliotis, ``Consistency
  analysis and improvement of vision-aided inertial navigation,'' \emph{IEEE
  Transactions on Robotics}, vol.~30, no.~1, pp. 158--176, 2014.

\bibitem{wu2017_vinsonwheels}
K.~J. Wu, C.~X. Guo, G.~Georgiou, and S.~I. Roumeliotis, ``{VINS} on wheels,''
  in \emph{IEEE ICRA}, 2017.

\bibitem{thrun2005probabilistic}
S.~Thrun, W.~Burgard, and D.~Fox, \emph{Probabilistic robotics}.\hskip 1em plus
  0.5em minus 0.4em\relax MIT Press, 2005.

\bibitem{lee2020_viwo_onlinecalib}
W.~Lee, K.~Eckenhoff, Y.~Yang, P.~Geneva, and G.~Huang,
  ``Visual-{Inertial}-{Wheel} {Odometry} with {Online} {Calibration},'' in
  \emph{{IEEE}/{RSJ} {International} {Conference} on {Intelligent} {Robots} and
  {Systems} ({IROS})}, 2020.

\bibitem{WuReport}
\BIBentryALTinterwordspacing
K.~J. Wu and S.~I. Roumeliotis, ``Unobservable directions of vins under special
  motions,'' Department of Computer Science, University of Minnesota, Tech.
  Rep., September 2016. [Online]. Available:
  \url{http://mars.cs.umn.edu/research/VINSodometry.php}
\BIBentrySTDinterwordspacing

\end{thebibliography}
\end{document}